%% file: anonymous-submission-latex-2026.tex
\documentclass[letterpaper]{article} 
\usepackage{aaai2026}  
\usepackage{times}  
\usepackage{helvet}  
\usepackage{courier}  
\usepackage[hyphens]{url}  
\usepackage{graphicx} 
\usepackage{natbib}  
\usepackage{caption} 
\usepackage{arydshln}
\usepackage{algorithm}
\usepackage{algorithmic}
\usepackage{amsmath}
\usepackage{amssymb}
\usepackage{colortbl}
\usepackage{xcolor}         
\usepackage{multirow}
\usepackage{lipsum}
\usepackage[colorlinks=true, linkcolor=black, citecolor=black, urlcolor=black]{hyperref}
\newcommand{\SEG}{\texttt{<SEG>}\ }

\definecolor{lightgray}{RGB}{220,220,220}
\definecolor{lightpurple}{RGB}{230,230,255}
\newlength\savewidth\newcommand\shline{\noalign{\global\savewidth\arrayrulewidth
  \global\arrayrulewidth 0.8pt}\hline\noalign{\global\arrayrulewidth\savewidth}}
\newcommand{\pub}[1]{\color{gray}{\tiny{[{#1}]}}}
\usepackage{newfloat}
\usepackage{listings}
\DeclareCaptionStyle{ruled}{labelfont=normalfont,labelsep=colon,strut=off} 
\floatstyle{ruled}
\newfloat{listing}{tb}{lst}{}
\floatname{listing}{Listing}
\nocopyright 

\title{Reinforcing Video Reasoning Segmentation to Think Before It Segments}
\author{
    Sitong Gong, Lu Zhang, Yunzhi Zhuge, Xu Jia, Pingping Zhang, Huchuan Lu 
}
\affiliations{
    IIAU Lab, Dalian University of Technology}
\usepackage{bibentry}

\begin{document}

\maketitle

\begin{figure*}
\vspace{-2mm}
    \centering
    \includegraphics[width=0.98\linewidth]{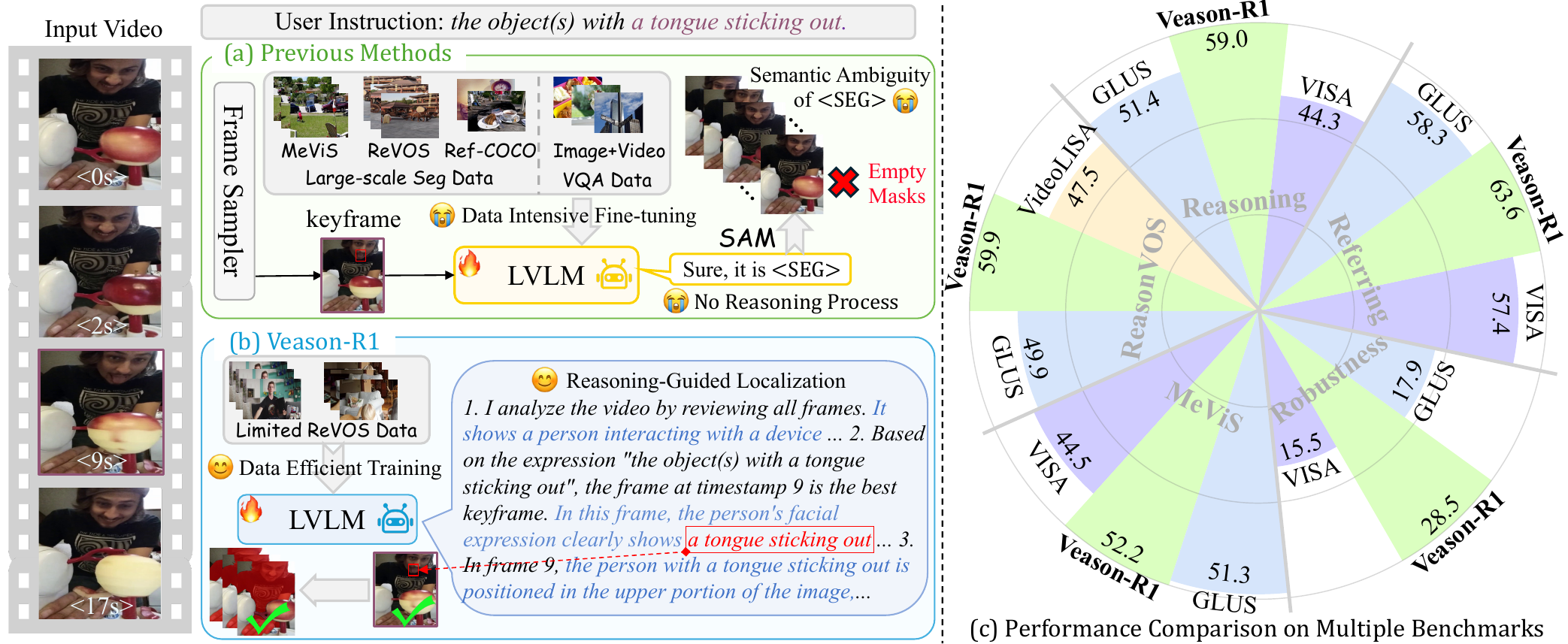}
    \vspace{-1mm}
    \caption{
    \textbf{Comparison of Veason-R1 with Existing Video Reasoning Segmentation (VRS) Approaches}. (a) Conventional methods fine-tune an LVLM using structured data, encoding global video-level information into a single \SEG token for segmentation guidance; however, this token exhibits weak reasoning capabilities, leading to misaligned masks. (b) In contrast, Veason-R1 introduces a reasoning-guided paradigm enhanced by spatiotemporal reinforcement learning, enabling step-by-step thinking to identify keyframes and produce interpretable segmentation with minimal training data. (c) Veason-R1 achieves SOTA performance on multiple benchmarks, demonstrating superior segmentation accuracy and hallucination robustness.
    }
    \label{fig:teasor}
\end{figure*}

\begin{abstract}
Video reasoning segmentation (VRS) endeavors to delineate referred objects in videos guided by implicit instructions that encapsulate human intent and temporal logic. Previous approaches leverage large vision language models (LVLMs) to encode object semantics into \SEG tokens for mask prediction. However, this paradigm suffers from limited interpretability during inference and suboptimal performance due to inadequate spatiotemporal reasoning. Drawing inspiration from seminal breakthroughs in reinforcement learning, we introduce Veason-R1, a specialized LVLM for VRS that emphasizes structured reasoning in segmentation. 
Veason-R1 is trained through Group Relative Policy Optimization (GRPO) augmented with Chain-of-Thought (CoT) initialization. To begin with, we curate high-quality CoT training data to instill structured reasoning trajectories, bridging video-level semantics and frame-level spatial grounding, yielding the supervised fine-tuned model Veason-SFT. Subsequently, GRPO fine-tuning encourages efficient exploration of the reasoning space by optimizing reasoning chains. To this end, we incorporate a holistic reward mechanism that synergistically enhances spatial alignment and temporal consistency, bolstering keyframe localization and fine-grained grounding. 
Comprehensive empirical evaluations demonstrate that Veason-R1 achieves state-of-the-art performance on multiple benchmarks, surpassing prior art by significant margins (\textit{e.g.}, +1.3 $\mathcal{J}\&\mathcal{F}$ in ReVOS and +10.0 $\mathcal{J}\&\mathcal{F}$ in ReasonVOS), while exhibiting robustness to hallucinations (+8.8 $\mathcal{R}$). Our code and model weights will be available at \href{https://github.com/SitongGong/Veason-R1}{Veason-R1}.

\end{abstract}

\section{1. Introduction}
Video reasoning segmentation (VRS)~\cite{yan2024visa,bai2024one} aims to produce pixel-wise mask sequences based on language queries encompassing human commonsense and implicit temporal logic.
Unlike traditional referring video object segmentation~\cite{seo2020urvos,wu2022language}, which depends on explicit descriptions (\textit{e.g.}, \textit{``the person on a skateboard''}), VRS harnesses world knowledge and temporal modeling in large vision language models (LVLMs)~\cite{liu2023visual,bai2025qwen2} to perceive complex dynamics with fine granularity.
The capability of modeling intricate temporal relations is critical for real-world applications that rely on sequential reasoning to 
support naunced perception and action, facilitating reliable decisions in domains such as robotic manipulation~\cite{billard2019trends,shridhar2022cliport} and autonomous driving~\cite{tian2024drivevlm,xie2025vlms}.

Representative approaches~\cite{yan2024visa,bai2024one,zheng2024villa,wei2024instructseg,gong2025devil} in VRS fields typically employ the LVLM to transform the language query into specialized tokens that serve as semantic embeddings of the referred targets across the video, followed by a mask decoder to produce the corresponding mask trajectories.
Despite achieving strong performance, these methods face two key limitations:
(i) \textit{\textbf{Limited Reasoning and Semantic Alignment.}} Prior approaches inject video-level information into segmentation tokens but lack structured reasoning traces, leading to inherent semantic ambiguity. As shown in Fig.~\ref{fig:teasor} (a), this diminishes efficacy in reasoning-intensive scenarios, such as long videos with temporal occlusions or evolving object interactions, where multi-step inference is essential.
(ii) \textit{\textbf{Reliance on Large-Scale Training Data.}} Token-based methods demand extensive annotated datasets for LVLM fine-tuning, as associating specialized tokens with image embeddings
requires diverse examples for cross-modal alignments and spatiotemporal handling (\textit{e.g.}, motion, occlusions).
VISA~\cite{yan2024visa} exemplifies this by training on a mixture of image and video segmentation datasets (encompassing 8.8k videos and 214k images), thereby inflating costs and impeding scalability, efficiency, and low-resource generalization.

Recent research~\cite{guo2025deepseek,shao2024deepseekmath} demonstrates that reinforcement learning (RL) fine-tuning elicits structured, interpretable reasoning from large language models (LLMs) at inference time. In particular, Group Relative Policy Optimization (GRPO) enhances in-context reasoning by estimating relative advantages within response groups, enabling critic-free optimization that is both efficient and data-sparse over traditional RL methods. Building on this, several works~\cite{liu2025seg,wang2025pixelthink,huang2025sam,you2025seg} have applied GRPO to reasoning-based segmentation tasks, using reward functions based on format constraints and IoU metrics to enhance segmentation quality. Meanwhile, others~\cite{feng2025video,li2025videochat,bai2025univg,zhong2025omni} have extended GRPO to video understanding and multi-image grounding, achieving more coherent reasoning and robust grounding in complex visual scenarios. Inspired by these advancements, we propose a RL framework for video reasoning segmentation that integrates Chain-of-Thought (CoT) imitation learning with GRPO-based fine-tuning.

To develop structured reasoning capabilities for identifying critical frames and improving grounding performance in dynamic visual contexts, we first curate a CoT dataset comprising 5.8k annotated samples. This dataset is used to perform supervised fine-tuning (SFT) of the Qwen2.5-VL model~\cite{bai2025qwen2}. Specifically, we employ customized prompt templates to guide Seed1.5-VL~\cite{guo2025seed1} in generating CoT reasoning traces, which serve as ground-truth supervision. These reasoning traces guide the model to analyze video content, identify keyframes from referring expressions, and perform accurate grounding within the identified frames. Through this imitation learning process, \textbf{Veason-SFT} acquires both keyframe analysis and basic object grounding capabilities.
To further hone these abilities, we fine-tune on 10k samples from the ReVOS dataset using GRPO. This stage incorporates a tailored reward policy that synergistically supervises spatial precision and temporal relevance. The reward policy includes a temporal localization reward that evaluates keyframe saliency, a spatial alignment reward that quantifies spatial grounding accuracy, and a unified consistency reward, which integrates SAM2~\cite{ravi2024sam} to fortify the coherence between keyframe selection and spatial grounding. Overall, this two-stage training pipeline yields robust, interpretable reasoning and precise visual grounding in complex videos, as illustrated in Fig.~\ref{fig:teasor} (b). The final \textbf{Veason-R1} model attains state-of-the-art performance across multiple VRS benchmarks, exhibiting superior segmentation accuracy and enhanced robustness against hallucinations, as shown in Fig.~\ref{fig:teasor} (c). 


Key contributions can be summarized as follows:
\begin{itemize}
    \item We introduce \textbf{Veason-R1}, the first approach to video reasoning segmentation that employs reinforcement learning. Specifically, we leverage GRPO-driven policy optimization, initialized with structured fine-tuning, to jointly enable keyframe identification and spatial grounding using only 10k training samples, a significant reduction compared to the 192k samples required by priors.
    \item We curate a chain-of-thought (CoT) dataset to equip the model with hierarchical reasoning capabilities, bridging video-level understanding and frame-level object grounding. Furthermore, we design a complementary reward policy during the GRPO stage to enhance temporally coherent reasoning and fine-grained localization.
    \item Experiments demonstrate that \textbf{Veason-R1} consistently obtains superior performance on the ReVOS, ReasonVOS and MeViS benchmarks, validating its effectiveness in logical reasoning and detailed visual comprehension.
\end{itemize}

\section{2. Related Works}


\textbf{Video Reasoning Segmentation (VRS)}
 is an emerging task requiring explicit multi-modal reasoning to segment target objects in video based on natural language queries~\cite{yan2024visa,bai2024one,zheng2024villa}.
 Pioneering work such as VISA~\cite{yan2024visa} combines keyframe sampling with a large vision language model (LVLM) for temporal reasoning and employs an object tracker for mask propagation. VideoLISA~\cite{bai2024one} introduces a sparse-to-dense sampling strategy and the One-Token-Seg-All paradigm, enabling video-level segmentation via a unified token representation. HyperSeg~\cite{wei2024hyperseg} generalizes the unified reasoning framework through a hybrid entity recognition mechanism, supporting cross-domain segmentation across both image and video inputs.
Further advances include VRS-HQ~\cite{gong2025devil}, enhancing temporal consistency through token fusion and occlusion-aware keyframe selection with SAM2, and Sa2VA~\cite{yuan2025sa2va}, integrating SAM2 with LVLM for balanced performance in multiple dense grounding and conversational tasks.

However, such segmentation token-based approaches are limited by the semantic ambiguity of the token representation, often misaligned with the actual objectives and inherently lack interpretability.
To address these limitations, we focus on leveraging the GRPO algorithm to enhance the implicit reasoning in VRS, jointly optimizing segmentation performance across spatial and temporal dimensions. 

\vspace{3mm}
\noindent \textbf{Visual Reinforcement Fine-Tuning.}
Reinforcement learning~\cite{sutton1998reinforcement}  has emerged as a powerful paradigm for optimizing large language models, particularly for enhancing their reasoning capabilities, as demonstrated by ChatGPT-o1\cite{jaech2024openai}.
Deepseek-R1~\cite{guo2025deepseek} introduces group relative policy optimization (GRPO),which leverages verifiable reward signals to estimate relative advantages among responses, thereby substantially improving reasoning.
Building on this, GRPO-based fine-tuning has been extended to various multimodal tasks, including image spatial reasoning~\cite{liu2025visual,liu2025seg,wang2025pixelthink}, video understanding~\cite{feng2025video,li2025videochat,wang2025time}, multi-image grounding~\cite{bai2025univg,zhang2025improving}, and visual generation~\cite{fang2025got,xiao2025mindomni,xue2025dancegrpo}, showcasing its versatility in challenging multimodal scenarios.
Earlier methods such as Seg-Zero~\cite{liu2025seg} and VisionReasoner~\cite{liu2025visionreasoner} incorporate segmentation-specific reward designs, leading to improved performance in image-level reasoning segmentation tasks.
Inspired by these, Omni-R1~\cite{zhong2025omni} applies GRPO to Ref-AVS and VRS; however, its cascaded dual-system architecture relies on large-scale training data and a pre-trained VRS model.
In contrast, our Veason-R1 leverages a single system that simultaneously performs keyframe localization and object grounding using only 10k fine-tuning samples, delivering superior segmentation accuracy in complex video scenarios. 

\section{3. Preliminary}
Group Relative Policy Optimization (GRPO)~\cite{guo2025deepseek} is a reinforcement learning algorithm derived from Proximal Policy Optimization (PPO)~\cite{schulman2017proximal}, eliminating the need for a separate value function by using relative rewards from grouped samples.
For a given prompt $p$, GRPO samples $G$ responses $o=\{o_1, ..., o_G\}$ from the policy $\pi_\theta$. A scalar reward function $r(\cdot)$ evaluates each response, yielding $\{r(o_1), ..., r(o_G)\}$. These rewards are then normalized within the group to compute relative advantages:

\begin{equation}
\label{Equation 1}
    A_{i} = \frac{r(o_{i})-mean(\{r(o_{i})\}^{G}_{i=1})}{std(\{r(o_{i})\}^{G}_{i=1})},
\end{equation}
where $A_{i}$ represents the relative advantage of the $i$-th candidate response. This encourages the model to favor higher-quality outputs within each sampled group, without relying on absolute reward calibration.
To stabilize training and avoid policy collapse, GRPO further introduces a KL-divergence regularization term $D_{KL}(\cdot||\cdot)$ to penalize deviations from a reference policy $\pi_{\text{ref}}$, resulting in the following optimization objective:
\begin{equation}
\label{Equation 2}
   \max_{\pi_{\theta}} \mathbb{E}_{o \sim\pi_{\theta_{\text{old}}}}(p) [ \sum^{G}_{i=1}(\frac{\pi_{\theta}(o_{i})}{\pi_{\theta_{\text{old}}(o_{i})}}) \cdot A_{i} - \beta{D_{KL}(\pi_{\theta}||\pi_{\text{ref}})} ],
\end{equation}
with $\beta$ serving as a hyperparameter that modulates the intensity of the regularization. This balanced objective encourages the policy model to generate responses with higher relative rewards while preserving proximity to the original policy distribution, supporting reliable convergence in reinforcement learning scenarios.

\begin{figure*}
    \centering
    \vspace{-3mm}
    \includegraphics[width=0.97\linewidth]{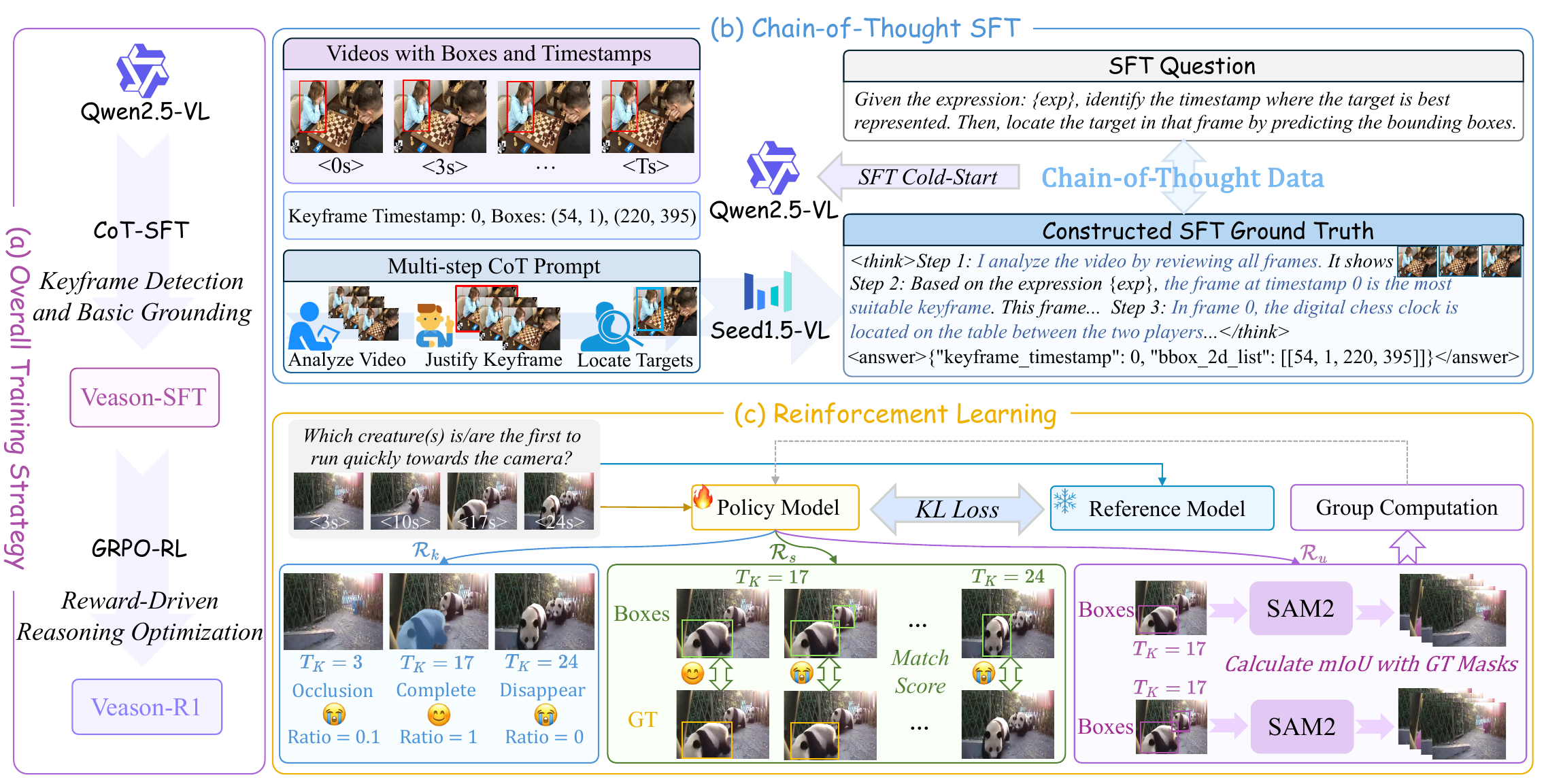}
    \vspace{-1mm}
    \caption{\textbf{Overall Training Pipeline of Veason-R1}. 
(a) Our two-stage training strategy comprises
\textbf{Veason-SFT} (CoT-based supervised fine-tuning (SFT) for reasoning-aware keyframe selection and basic grounding), followed by \textbf{Veason-R1} (reinforcement learning (RL) via GRPO to enhance reasoning fidelity).
(b) In the SFT stage, multi-step CoT prompts and pseudo annotations guide Seed1.5-VL to generate structured reasoning traces, which are combined with annotations for LVLM supervision.
(c) During RL, the policy model samples candidate keyframes and box predictions. 
The reward policy jointly reinforces spatial alignment of boxes and temporal consistency in keyframe selection, with KL-regularization ensuring training stability. 
}
    \label{fig:pipeline}
\end{figure*}

\section{4. Methodology}
\subsection{Overview}

Given a video sequence $\mathcal{V}_{I} \in \mathbb{R}^{T \times 3 \times H \times W}$ consisting of $T$ RGB frames, where $H$ and $W$ denote the height and width of each frame, and a reasoning instruction $Q_{\text{txt}}$, the goal of VRS is to generate a sequence of segmentation masks $\mathcal{M} \in \mathbb{R}^{T \times H \times W}$ using the model.

Unlike prior VRS approaches~\cite{yan2024visa,bai2024one,zheng2024villa} that embed object semantics into specialized tokens without offering interpretability or transparency, our method explicitly encourages step-by-step reasoning prior to mask generation. We formulate the task as a two-stage process: the model first analyzes the video $\mathcal{V}_{I}$ in conjunction with the referring expression $Q_{\text{txt}}$ to identify a keyframe timestamp $T_{k}$ in which the referred object is most representative. It then performs spatial grounding by predicting a set of bounding boxes $\mathcal{B}_{T_{k}} = \{b_{i}\}_{i=1}^{N_{k}}$, where $N_{k}$ is the number of detected instances in the keyframe and each $b_{i}$ corresponds to a bounding box in $(x_1, y_1, x_2, y_2)$ format. This explicit decomposition enhances both interpretability and spatiotemporal grounding accuracy.

As illustrated in Fig.~\ref{fig:pipeline} (a), we adopt a two-stage training paradigm. We initialize our framework with Qwen2.5-VL~\cite{bai2025qwen2} as the LVLM. In the first stage, we construct a chain-of-thought (CoT) dataset and perform supervised fine-tuning, resulting in the Veason-SFT model. This stage equips the model with hierarchical video reasoning capabilities, enabling it to identify keyframes and perform coarse object localization. In the second stage, we apply GRPO to refine the model's reasoning space, producing the Veason-R1 model with improved spatiotemporal grounding and enhanced reasoning coherence.


\subsection{CoT-based Supervised Fine-Tuning}
Directly applying reinforcement learning (RL) to advanced LVLMs for VRS often leads to unstable training and suboptimal reasoning behavior. This limitation stems from the inability of Qwen2.5-VL to perform structured reasoning under complex temporal dynamics and implicit queries, which hinders effective exploration of reasoning trajectories during RL optimization.
To address this challenge, we first construct a high-quality CoT dataset to provide structured cold-start supervision. This initialization pre-equips the model with task-specific reasoning priors and guides it to form hierarchical understanding across video-level semantics and frame-level spatial details, which serves as a stable foundation for the subsequent RL fine-tuning.

\noindent\textbf{CoT Data Construction.}
To construct instructive CoT reasoning traces, we design a data generation pipeline to guide Seed1.5-VL~\cite{guo2025seed1} in generating step-by-step reasoning processes, as illustrated in Fig.~\ref{fig:pipeline} (b). This approach ensures accurate generated reasoning traces while reducing manual annotation costs. 
Specifically, we randomly sample one pseudo keyframe from the top-5 frames with the largest target area in each video to prevent the model from overfitting to fixed keyframe indices. 
Based on this pre-selected frame, we devise a multi-step CoT prompt that instructs the model to (i) analyze the scene, (ii) substantiate the frame's relevance to the expression, and (iii) localize the referred objects within this frame. 
Subsequently, we delineate all referred objects using bounding boxes in sampled video frames, and feed the annotations, timestamps, and the predefined CoT prompt into Seed1.5-VL for generating reasoning chains. 
We encapsulate the final CoT reasoning process within \texttt{<think> </think>} tags, and place the selected keyframe and bounding box coordinates inside \texttt{<answer> </answer>} tags to serve as the ground truth. 

\noindent\textbf{Supervised Fine-Tuning.}
Leveraging our constructed CoT dataset, we fine-tune the Qwen2.5-VL model to endow it with both keyframe selection and preliminary spatial grounding capabilities. 
During training, we adopt the auto-regressive next-token prediction paradigm, treating the thinking process and final answer as a unified text sequence and applying cross-entropy loss over the entire generation sequence. 
Given the limited size of the CoT dataset, we adopt LoRA-based~\cite{hu2022lora} fine-tuning to efficiently enhance reasoning capabilities while mitigating overfitting. 

\subsection{Reward-Driven Reasoning Optimization via GRPO}
Building upon Veason-SFT's capabilities in keyframe selection and basic grounding, we adopt a GRPO-based RL strategy to further enhance the model's contextual reasoning quality.
GRPO uses preference-based rewards to guide the model in generating reasoning chains that more precisely focus on the salient frame and referred targets, enabling it to handle intricate temporal dynamics and generalize across diverse video scenarios. 
As illustrated in Fig.~\ref{fig:pipeline} (c), the policy model explores the action space by generating $G$ candidate outputs per iteration and receives feedback through multiple task-aligned reward signals.
Following the GRPO framework, we estimate relative advantages from sampled outputs (\textit{cf} Eq.~\ref{Equation 1}) and update the policy model accordingly (\textit{cf}  Eq.~\ref{Equation 2}), with KL regularization for stable optimization. 

Given the pivotal role of the reward function in effective policy optimization, we devise a tailored reward mechanism that evaluates reasoning quality from key perspectives, incorporating a temporal localization reward for keyframe saliency, a spatial alignment reward for detection accuracy, and a unified consistency reward for spatiotemporal coherence between selected frames and grounded objects.

\noindent\textbf{Format Compliance Reward.} We employ the reward $\mathcal{R}_{f}$ to incentivize a structured, step-by-step thinking process and enforce strict output formatting: the reasoning trace must be enclosed within \texttt{<think> </think>} tags, while the final answer must appear within \texttt{<answer> </answer>} tags, formatted as a dictionary containing two mandatory fields, \textit{i.e.}, keyframe\_timestamp and bbox\_2d\_list.

\noindent\textbf{Temporal Localization Reward.} 
The reward $\mathcal{R}_{k}$ encourages the model to select frames in which the referred object is most visually prominent. Given $\hat{T}$ uniformly sampled video frames $\mathcal{V}_{s}$, we prepend each image token with their corresponding timestamp (\textit{e.g.}, \texttt{<0s>}). For the predicted keyframe timestamp $T_k$, we compute the reward as the ratio between the mask area of the referred object in frame $T_k$ and the maximum mask area across all sampled frames:
\begin{equation}
    \mathcal{R}_{k} = \frac{\mathcal{S}_{T_{k}}}{max(\{\mathcal{S}_{t}\}^{\hat{T}}_{t=1})},
\end{equation}
where $\mathcal{S}_{t}$ represents the foreground area of the frame at $t$. 

\noindent\textbf{Spatial Alignment Reward.} 
The reward $\mathcal{R}_{s}$ measures the accuracy of object localization in the predicted keyframe. Given the predicted bounding boxes $\mathcal{B}_{T_k} = \{b_i\}^{N_{k}}_{i=1}$ and the ground-truth boxes $\mathcal{B}^{gt}_{T_k} = \{b^{gt}_j\}^{N^{gt}_{k}}_{j=1}$ at keyframe $T_k$, where $N_{k}$ and $N^{gt}_{k}$ refer to the number of predicted and ground-truth targets within keyframe, we compute a pairwise IoU-based cost matrix $C \in \mathbb{R}^{N_{k} \times N^{gt}_{k}}$ as:
\begin{equation}
C_{i,j} = 1 - IoU(b_i, b^{gt}_j),
\end{equation}
where $IoU(\cdot)$ denotes the intersection-over-union between two bounding boxes.
To support multi-object grounding and enforce one-to-one matching, we apply the Hungarian algorithm to obtain the optimal assignment $(i^*, j^*)$. The final reward is computed as the normalized sum of matched IoUs:
\begin{equation}
    \mathcal{R}_{s} = \frac{1}{\max(N_{k}, N^{gt}_{k})} \sum_{(i,j)\in {C'}} IoU(b_{i}, b^{gt}_{j}),
\end{equation}
where $C'$ denotes the set of matched index pairs derived from the Hungarian algorithm. 
This formulation effectively promotes precise spatial alignment, even in scenarios involving multiple referred targets.

\noindent\textbf{Unified Consistency Reward.} 
To jointly evaluate the temporal consistency of keyframe selection and the spatial alignment of predicted boxes, we introduce the reward $\mathcal{R}_{u}$. After performing Hungarian matching on the predicted boxes $\mathcal{B}_{T_k}$ in the selected keyframe, we obtain a set of matched boxes $\mathcal{B}'_{T_k}$ aligned with the ground-truths. These matched boxes, along with the keyframe timestamp, are fed into a frozen SAM2~\cite{ravi2024sam} to generate video-level segmentation masks $\mathcal{M}=\{m_{t}\}^{\hat{T}}_{t=1}$: 
\begin{equation}
  \mathcal{M} = \text{SAM2}(\mathcal{B}'_{T_k}, \mathcal{V}_{s}, T_k)
\end{equation}
Subsequently, we merge all predicted masks across objects and compute the average IoU with the ground-truth masks $\mathcal{M}^{gt}=\{m^{gt}_{t}\}^{\hat{T}}_{t=1}$ across sampled frames:
\begin{equation}
\mathcal{R}_{u}=\frac{1}{\hat{T}} \sum^{\hat{T}}_{t=1}IoU(m_{t},m^{gt}_{t})
\end{equation}
The total reward $\mathcal{R}_{total}$ is computed as a weighted sum of the four sub-rewards described above:
\begin{equation}
    \mathcal{R}_{total} = \alpha_{f}\mathcal{R}_{f} + \alpha_{k}\mathcal{R}_{k} + \alpha_{s}\mathcal{R}_{s} + \alpha_{u}\mathcal{R}_{u},
\end{equation}
where the coefficients $\alpha_{f}$, $\alpha_{k}$, $\alpha_{s}$, and $\alpha_{u}$ are all set to 1.0.




\section{5. Experiments}

\input{Tables/revos_comparison}
\input{Tables/reason_mevis_results_}

\subsection{Datasets and Metrics}
We select training samples from the ReVOS dataset, a large-scale VRS benchmark with diverse and complex scenarios, which are crucial for evaluating temporal reasoning and spatial grounding capabilities.  
In the supervised fine-tuning (SFT) stage, we adapt the Qwen2.5-VL using our curated chain-of-thought dataset tailored for video reasoning segmentation task. During group relative policy optimization (GRPO) training, we sample 10,000 instances proportional to the referring and reasoning subsets in the ReVOS dataset. For evaluation, we rigorously assess Veason-R1 across diverse benchmarks, including the validation set of ReVOS, ReasonVOS~\cite{bai2024one} that features longer videos and more implicit expressions, and MeViS~\cite{ding2023mevis} that focuses on multi-object and motion-intensive expressions.
Consistent with prior works~\cite{yan2024visa}, we adopt region similarity $\mathcal{J}$, contour accuracy $\mathcal{F}$, and their average $\mathcal{J}\&\mathcal{F}$ as primary metrics. 
Specifically, $\mathcal{J}$ quantifies the intersection-over-union (IoU) of predicted and ground-truth mask sequence, whereas $\mathcal{F}$ measures the contour-based alignment. In addition, we adopt the robustness score $\mathcal{R}$ to evaluate the model's resistance to hallucination.

\subsection{Implementation Details}
In the supervised fine-tuning (SFT) stage, we use LLaMA-Factory~\cite{zheng2024llamafactory} to fine-tune the Qwen2.5-VL with LoRA~\cite{hu2022lora} (rank 8), freezing other parameters. We apply a learning rate of $1 \times 10^{-4}$, cosine annealing, and gradient accumulation of 8 steps, training for one epoch on our video-tailored CoT dataset.
In the reinforcement learning (RL) stage, we employ the VERL~\cite{sheng2025hybridflow} framework with a global batch size of 16, sampling 8 responses per input prompt to facilitate preference optimization, and a learning rate of $1 \times 10^{-6}$ for one epoch.  
Experiments are conducted on two NVIDIA A100 GPUs. 

\subsection{Quantitative Comparison}
\label{ReVOS Evaluation}

\noindent\textbf{ReVOS.}
\label{ReVOS Evaluation}
Tab.~\ref{ReVOS} presents a detailed comparison between Veason-R1 and previous VRS approaches~\cite{yan2024visa,wei2024instructseg,wei2024hyperseg,gong2025devil}.
Notably, despite being fine-tuned on merely 10k samples, Veason-R1-3B attains performance commensurate with the prior state-of-the-art VRS-HQ-13B, while Veason-R1-7B surpasses it by 1.3 in $\mathcal{J}\&\mathcal{F}$. 
Particularly on the reasoning subset, Veason-R1-7B realizes an improvement of 2.2 in $\mathcal{J}\&\mathcal{F}$, underscoring the efficacy of integrating structured reasoning for guiding precise segmentation under complex temporal dynamics. 
Furthermore, Veason-R1 manifests a substantially higher robustness score $\mathcal{R}$ than prior methods, indicating that promoting structured reasoning prior to segmentation helps reduce hallucinations and enhance prediction reliability. 

\noindent\textbf{ReasonVOS. }
\label{ReasonVOS Evaluation}
Tab.~\ref{ReasonVOS} compares the segmentation performance of Veason-R1 against prior methods~\cite{bai2024one,lin2025glus} on the ReasonVOS benchmark, which comprises 91 videos with complex scenes (averaging 105 frames each) and 253 long-form  queries. These instructions include cases involving causal reasoning (\textit{e.g.}, inferring intent from behavior) and hypothetical scenarios (\textit{e.g.}, conditional or counterfactual prompts). Veason-R1 markedly outperforms GLUS by 8.5 in $\mathcal{J}$, 11.4 in $\mathcal{F}$, and 10.0 in $\mathcal{J}\&\mathcal{F}$, demonstrating its superior ability to handle intricate linguistic reasoning and temporally extended, dynamic video content.

\noindent\textbf{MeViS.}
\label{MeViS Evaluation}
In addition to VRS benchmarks, we further assess the generalization ability of Veason-R1 on the MeViS dataset, as illustrated in Tab.~\ref{mevis}. 
In contrast to existing methods that incorporate MeViS into their training data, Veason-R1 is trained solely on 10k samples from ReVOS and evaluated on MeViS in a zero-shot setting. 
Despite this discrepancy, Veason-R1 still surpasses previous art by 0.9 in $\mathcal{J}\&\mathcal{F}$, demonstrating the robustness and adaptability of our ``thinking before segmenting'' paradigm in tackling motion-centric and referring-based queries without task-specific tuning.

\subsection{Qualitative Comparison}
\begin{figure*}[!htbp]
    \centering
    \vspace{-3mm}
    \includegraphics[width=0.95\linewidth]{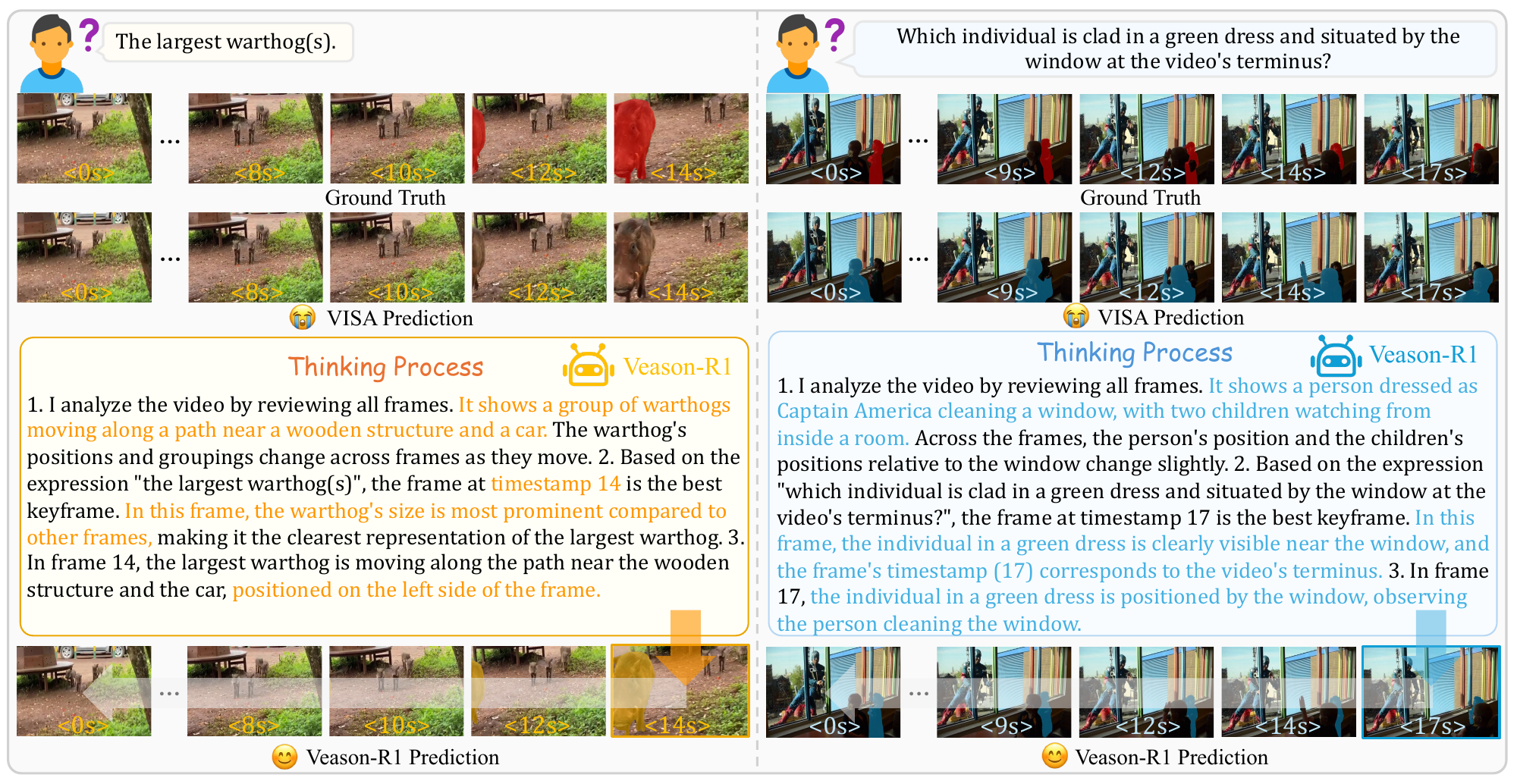}
    \vspace{-1mm}
    \caption{\textbf{Qualitative Comparison between Veason-R1-3B with VISA-7B}. 
    In challenging scenarios where the target disappears for a long duration (left) or the expression is highly coupled with temporal context (right), Veason-R1 generates a clear and interpretable reasoning process, accurately selects the keyframe, and reliably segments the semantically referred object.
    }
    \label{fig:qualitative comparison}
\end{figure*}

Fig.~\ref{fig:qualitative comparison} qualitatively compares Veason-R1-3B with VISA-7B in challenging scenarios involving occlusions and temporally grounded expressions.
For the left example, where the target emerges only in the final frames, Veason-R1 uses temporal reasoning to pinpoint the frame where the warthog is most prominent and accurately localizes it, whereas VISA fails to detect and outputs an empty mask sequence.
In the right case, Veason-R1 accurately interprets the visual scene and query to locate the girl ``situated by the window'' at the video end. These results underscore Veason-R1's strength in handling temporally sensitive and spatially complex instructions via explicit step-by-step reasoning. Additional qualitative results are available in the \textit{Supplementary Materials}.

\input{Tables/reward_cotsft_ablation}
\input{Tables/keyframe_ablation}

\subsection{Ablation Study}
We perform extensive ablation study based on Qwen2.5-VL-3B to assess the efficacy of the proposed reward policy, training strategy, and keyframe-first grounding strategy.

\noindent\textbf{Reward Policy.}
The results of ablating components of the reward functions are presented in Tab.~\ref{reward ablation}. 
Removing the $\mathcal{R}_{s}$ induces the most substantial performance decline (2.0 and 2.9 in $\mathcal{J}\&\mathcal{F}$ on referring and reasoning subsets), underscoring the critical role of frame-level spatial alignment in enhancing the model's localization capability.
Furthermore, omitting $\mathcal{R}_{u}$ yields performance degradations of 0.6 and 0.7 in $\mathcal{J}\&\mathcal{F}$ for the two subsets, demonstrating the effectiveness of jointly optimizing keyframe selection and spatial localization to maintain their temporal coherence. 

\noindent\textbf{Training Strategy.}
Tab.~\ref{cotsft ablation} presents the performance comparison across various fine-tuning strategies. 
We first assess the impact of incorporating the chain-of-thought (CoT) process during the SFT stage. Removing CoT reasoning leads to substantial performance drops of 12.8 and 8.5 in $\mathcal{J}\&\mathcal{F}$ for the referring and reasoning subsets, respectively. Even after subsequent GRPO fine-tuning, the model still lags behind Veason-R1 by 1.4 and 0.8 in $\mathcal{J}\&\mathcal{F}$, underscoring the effectiveness of explicitly supervising the reasoning process for improving both reasoning quality and grounding accuracy.
Next, we evaluate the effect of training with SFT and GRPO separately. We find that applying GRPO alone brings notable gains, outperforming CoT-SFT by 8.9 and 12.6 in $\mathcal{J}\&\mathcal{F}$. However, combining CoT-SFT with GRPO yields the best overall results, surpassing GRPO-only by 3.0 (referring) and 2.6 (reasoning) in $\mathcal{J}\&\mathcal{F}$, indicating the complementary benefits of initializing the model with structured reasoning chains before preference optimization.

\noindent\textbf{Keyframe Selection and Grounding.}
We investigate the impact of jointly modeling keyframe selection and grounding as shown in Tab.~\ref{keyframe ablation}. 
Grounding-only indicates directly applying GRPO with the first target-containing frame fixed as the keyframe, while SFT-keyframe denotes training the model exclusively for keyframe detection during the SFT stage.
Directly applying GRPO with a fixed keyframe significantly decreases performance by 4.7 and 5.5 in $\mathcal{J}\&\mathcal{F}$ for the referring and reasoning subsets, highlighting the critical need for accurate temporal localization prior to spatial grounding.
Furthermore, training the model to perform keyframe selection alone during SFT, while ignoring spatial grounding results in 
performance degradations of 8.4 and 5.8 in $\mathcal{J}\&\mathcal{F}$, underscoring the importance of jointly modeling temporal and spatial reasoning for optimal segmentation. 

\section{6. Conclusion}
We present Veason-R1, a reinforcement learning framework for VRS that explicitly models interpretable reasoning trajectories by decomposing the task into keyframe selection and object localization.
A two-stage training strategy is employed: first, a CoT-based SFT procedure instills hierarchical reasoning capabilities. Then, GRPO-based RL further refines reasoning and grounding behavior, guided by the reward policy that comprehensively captures spatial alignment, keyframe saliency, and temporal consistency. Veason-R1 achieves strong performance on multiple benchmarks, underscoring the effectiveness of structured pre-segmentation reasoning in enhancing multi-modal understanding of semantic and motion cues over time.

\appendix

\bibliography{aaai2026}

\section{Appendix}
In this supplementary material, we provide additional experimental details and qualitative analyses for Veason-R1. To begin with, we elaborate on the construction of the chain-of-thought training data (Section A). We then provide further implementation details to enhance reproducibility (Section B). Next, we analyze the training curves to illustrate the model's optimization dynamics and convergence behavior (Section C). This is followed by a discussion of representative failure cases to identify potential areas for improvement (Section D). Finally, we include additional qualitative visualizations to further demonstrate the effectiveness of Veason-R1 (Section E).

\subsection{A. More Details of CoT Training Data Construction}
\label{CoT Data}

\begin{figure*}[t]
    \centering
    \includegraphics[width=1\linewidth]{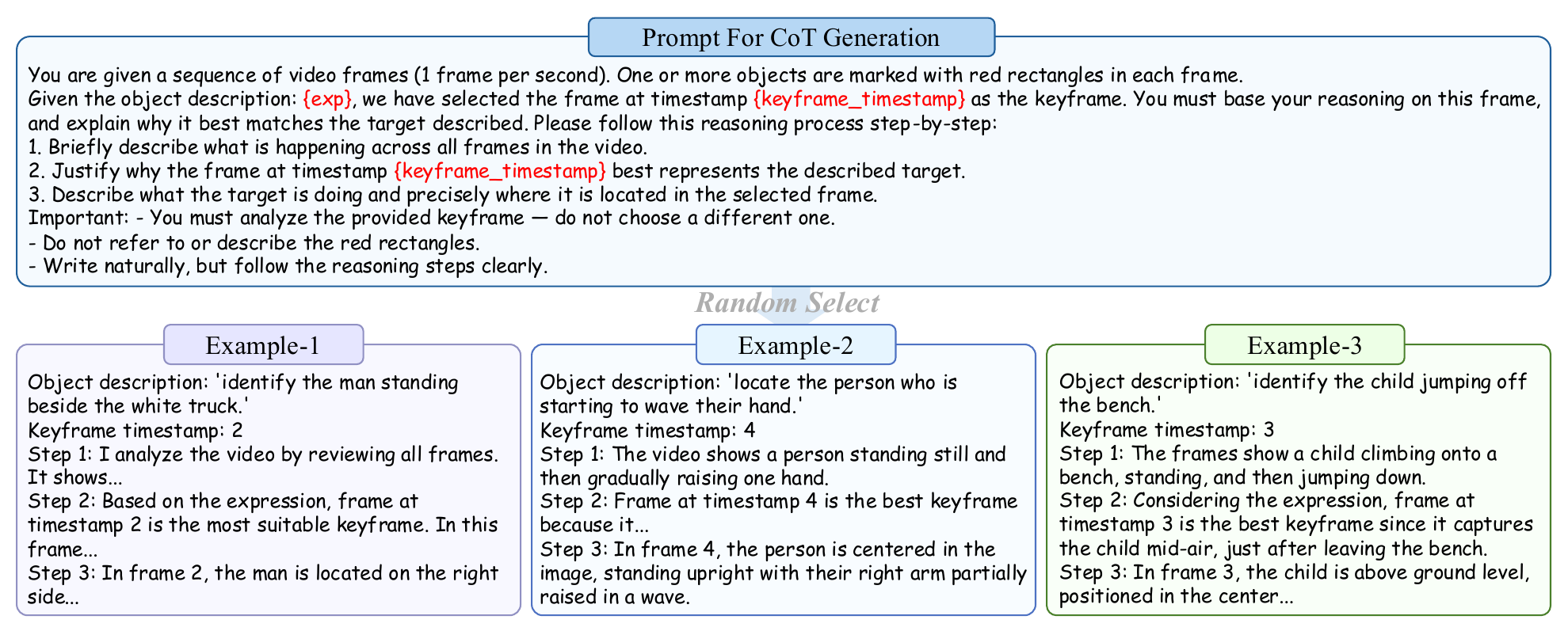}
    \caption{Prompt template for CoT data generation. }
    \label{fig:supple_cot_data}
\end{figure*}

\begin{figure*}[t]
    \centering
    \includegraphics[width=1\linewidth]{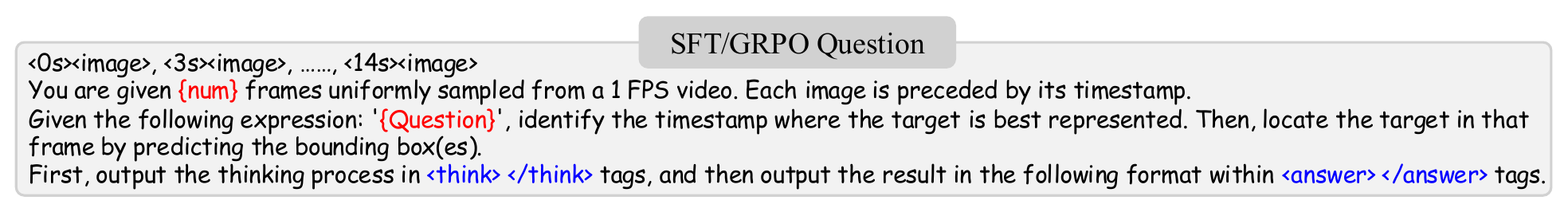}
    \caption{Question template of SFT and GRPO training stage. }
    \label{fig:supple_question}
\end{figure*}

\begin{figure*}[t]
    \centering
    \includegraphics[width=1\linewidth]{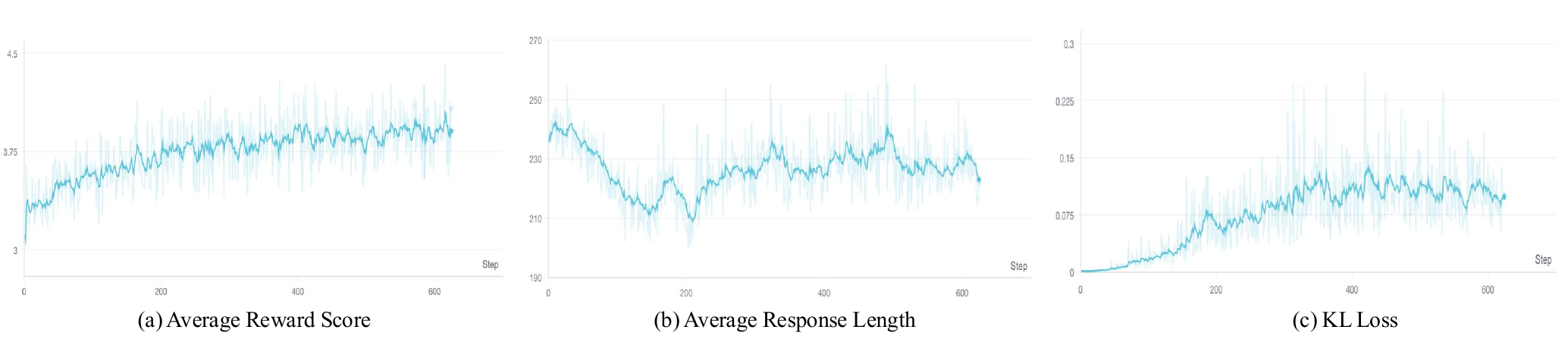}
    \caption{GRPO training curves. }
    \label{fig:supple_training_curves}
\end{figure*}

\begin{figure*}[t]
    \centering
    \includegraphics[width=1\linewidth]{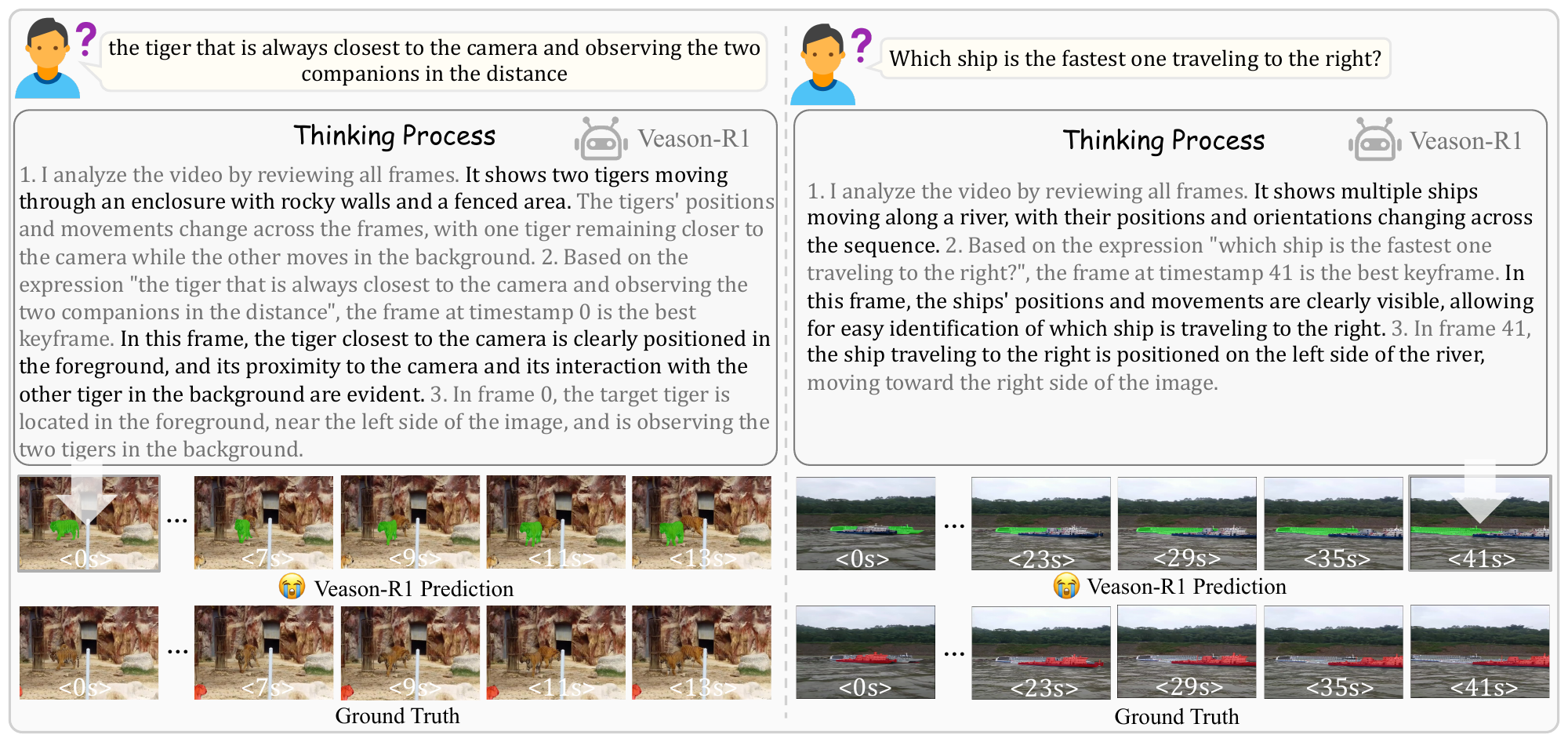}
    \caption{Visualization of failure cases on the ReVOS dataset. Veason-R1 exhibits inconsistencies between its reasoning traces and final segmentation outputs, and shows limited sensitivity to motion-related concepts such as object speed. }
    \label{fig:supple_failure_case}
\end{figure*}

\begin{figure*}[t]
    \centering
    \includegraphics[width=1\linewidth]{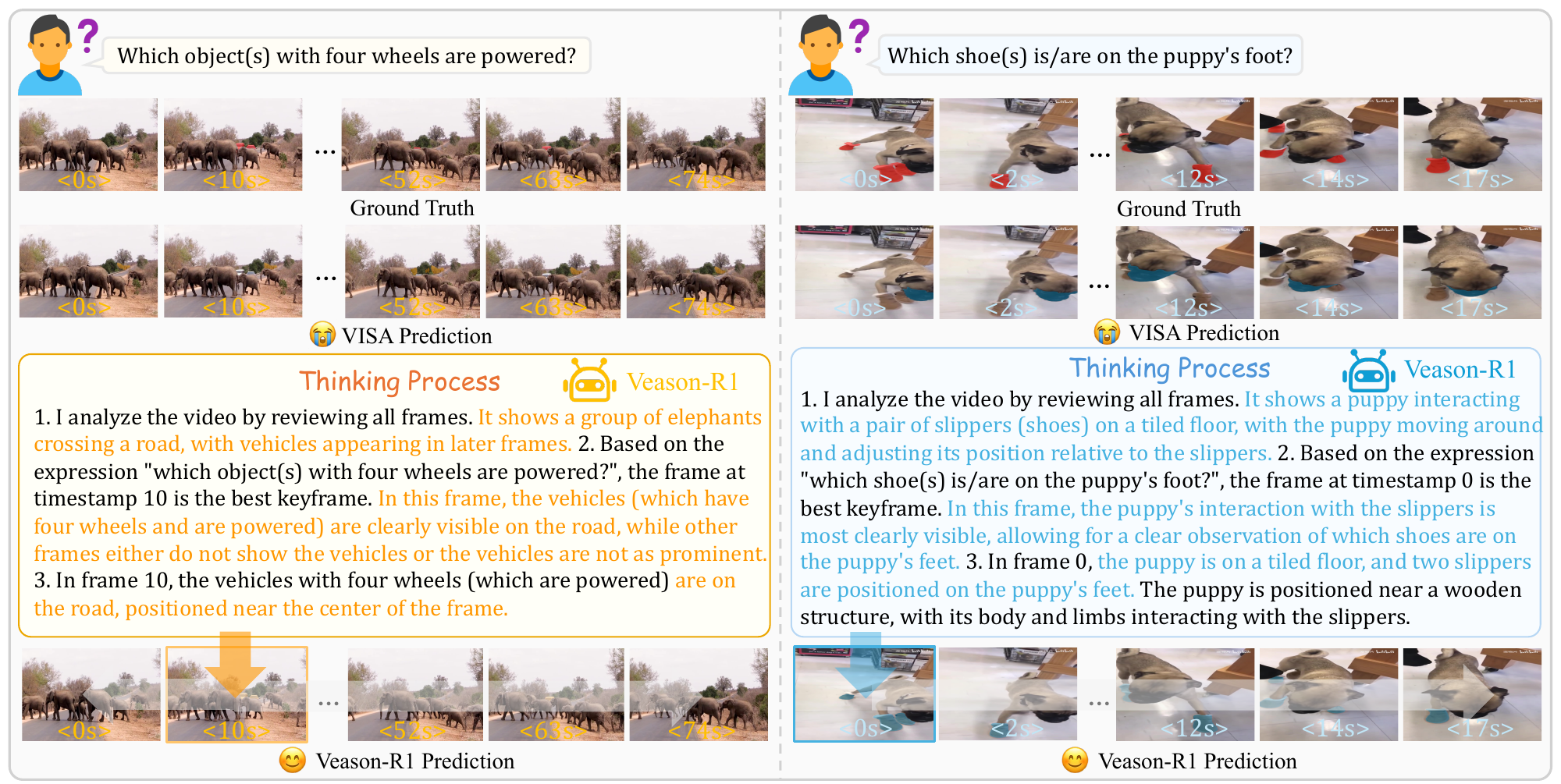}
    \caption{More visualization on the ReVOS dataset. }
    \label{fig:supple_visualization_revos}
\end{figure*}

\begin{figure*}[t]
    \centering
    \includegraphics[width=1\linewidth]{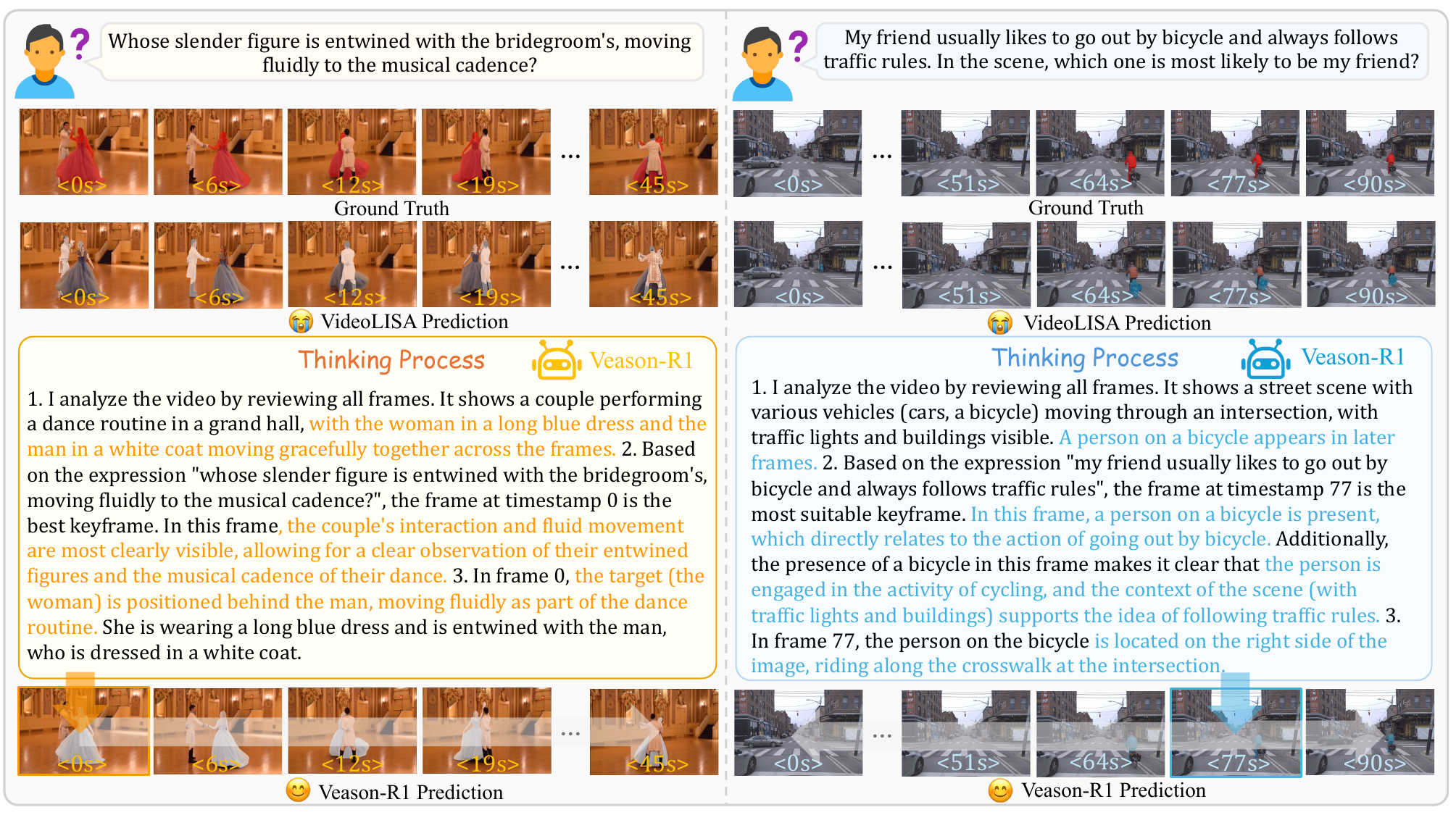}
    \vspace{-6mm}
    \caption{More visualization on the ReasonVOS dataset. }
    \label{fig:supple_visualization_reason_vos}
\end{figure*}

Fig.~\ref{fig:supple_cot_data} illustrates the full prompt that we use as input to Seed1.5-VL~\cite{guo2025seed1} for chain-of-thought generation.
The prompt is presented to the model along with a sequence of uniformly sampled video frames and their corresponding timestamps, as well as the predefined object description and a selected keyframe. 
The prompt guides the model to (i) briefly summarize the overall video content, (ii) justify why the selected keyframe best matches the target expression, and (iii) describe the target's actions and location within the given frame. 
This structured prompting encourages the model to produce clear and interpretable reasoning based on the visual and contextual information.
We observe that Seed1.5-VL tends to closely mimic the style of human-provided exemplars. To promote diversity in the generated CoT responses and avoid overly rigid reasoning patterns, we design three distinct CoT templates—each with slightly different phrasing and logical flow. For each sample, we randomly select one template during prompting, encouraging the model to explore different reasoning trajectories while maintaining consistency with the underlying task structure.

\subsection{B. More Implementation Details}
We adopt the same question template for both the SFT and GRPO stages, as illustrated in Fig.~\ref{fig:supple_question}. 
To improve keyframe grounding, we prepend real timestamps (e.g., \texttt{<0s>, <3s>}) before each sampled frame's image tokens, enabling the model to reason with accurate temporal alignment.
During the GRPO stage, we set the maximum response length to 1024 tokens, the maximum gradient norm for clipping to 1.0, and the coefficient for the KL divergence regularization loss to 5e-3. 
All input frames in both training stages are resized to a resolution of 560 $\times$ 560 pixels. 
During inference, we prompt Veason-R1 to generate the keyframe timestamp and the bounding boxes of the referred objects. These predictions are then passed to SAM2~\cite{ravi2024sam}, which performs segmentation and mask propagation across the entire video sequence, yielding the final predicted mask sequence.

\subsection{C. Training Curve Analysis}
We visualize the training dynamics of Veason-R1 in Fig.~\ref{fig:supple_training_curves}. We monitor three key metrics during training: total reward score $\mathcal{R}_{total}$, response length, and the value of KL loss. 

As shown in Fig.~\ref{fig:supple_training_curves} (a), the model initialized via CoT-based supervision starts with a reward score above 3.0, indicating a reasonable ability to generate structured outputs and locate keyframes. The consistent upward trend in reward reflects effective policy refinement and better alignment with the designed reward function.

It can be observed in Fig.~\ref{fig:supple_training_curves} (b) that the average response length initially decreases, suggesting that the model rapidly learns to eliminate redundant tokens and generate concise reasoning. In the later phase, the length slightly increases and then stabilizes, indicating an adjustment towards producing responses that balance brevity and informativeness.

As depicted in Fig.~\ref{fig:supple_training_curves} (c), the KL divergence gradually increases as the model departs from the initial supervised policy to explore more reward-driven behaviors. It eventually stabilizes, indicating convergence to a consistent and reward-aligned policy under the regularization constraint.

\subsection{D. Failure Case Analysis}
Fig.~\ref{fig:supple_failure_case} presents some failure cases of Veason-R1 on the ReVOS dataset. 
In the left example, the model generates a correct reasoning trace, accurately identifying the tiger closest to the camera and describing its relation to the background tigers. However, the predicted masks fail to align with the described target, revealing a disconnect between reasoning and visual grounding. This suggests that Veason-R1 struggles to translate reasoning into precise spatial localization, especially in cluttered scenes with visually similar objects.
In the right example, Veason-R1 correctly identifies the keyframe showing clear motion cues but ultimately selects the slower ship. The reasoning trace overlooks the constraint “fastest” and focuses solely on movement direction. This indicates the model’s limited sensitivity to speed and motion-related concepts, revealing weaknesses in capturing temporal dynamics or comparing relative movements. These limitations suggest the importance of incorporating data that better emphasizes motion perception in future work.

\subsection{E. More Visualization Results}
We present additional visual comparisons of Veason-R1-3B with VISA-7B~\cite{yan2024visa} and VideoLISA-3.8B~\cite{bai2024one} to underscore its robust reasoning capabilities and enhanced fine-grained grounding performance. 

The qualitative comparison of Veason-R1 with VISA is shown in Fig.~\ref{fig:supple_visualization_revos}. In the left example, a herd of elephants partially occludes the vehicles crossing the road. Veason-R1 successfully identifies the timestamp where both vehicles are most clearly visible and segments them, whereas VISA mistakenly highlights background trees. This reflects Veason-R1's robustness to occlusions and its focus on task-relevant semantics. The right case indicates that Veason-R1 correctly segments the slippers in contact with the puppy’s feet, demonstrating strong multi-object localization and spatial reasoning. VISA, by contrast, produces incorrect masks, revealing limitations in handling fine-grained interactions.

Fig.~\ref{fig:supple_visualization_reason_vos} visualizes the segmentation maps of Veason-R1 and VideoLISA on the ReasonVOS dataset. In the left example, Veason-R1 accurately segments the dancing bride in response to a poetic query involving abstract movement descriptions, whereas VideoLISA mistakenly segments the groom. This highlights Veason-R1's superior understanding of implicit and abstract language. In the right example, under a long-duration and high-complexity scenario, Veason-R1 successfully localizes the man riding the bicycle in response to a causal reasoning query, while VideoLISA yields fragmented and inconsistent predictions, demonstrating Veason-R1's stronger capacity for temporal reasoning and grounding in complex visual contexts.

\end{document}

%% file: Tables/revos_comparison.tex
\begin{table*}[!tbp]
\centering
\vspace{-3mm}
\caption{\textbf{Comparative Analysis on the ReVOS Benchmark.}
Detailed comparison of Veason-R1's performance against existing approaches on the ReVOS benchmark.
}
\label{ReVOS}
\footnotesize
\vspace{-1mm}
\scalebox{1.0}{
\begin{tabular}{rl||ccc|ccc|ccc|c}
\shline
&  & \multicolumn{3}{c|}{Referring} & \multicolumn{3}{c|}{Reasoning} & \multicolumn{3}{c|}{Overall} & \multirow{2}{*}{$\mathcal{R}$} \\
\multicolumn{2}{c||}{\multirow{-2}{*}{Methods}}  & $\mathcal{J}$ & $\mathcal{F}$ & $\mathcal{J}\&\mathcal{F}$ & $\mathcal{J}$ & $\mathcal{F}$ & $\mathcal{J}\&\mathcal{F}$ & $\mathcal{J}$ & $\mathcal{F}$ & $\mathcal{J}\&\mathcal{F}$ & \\
\shline
  \multicolumn{1}{r}{LMPM$_{\!}$~\cite{ding2023mevis}}&\!\!\!\pub{ICCV2023}\!\! & 29.0 & 39.1 & 34.1 & 13.3 & 24.3 & 18.8 & 21.2 & 31.7 & 26.4 & 3.2 \\
  \multicolumn{1}{r}{LISA-7B$_{\!}$~\cite{lai2024lisa}}& \!\!\!\pub{CVPR2024}\!\!  & 44.3 & 47.1 & 45.7 & 33.8 & 38.4 & 36.1 & 39.1 & 42.7  & 40.9 & 9.3\\
  \multicolumn{1}{r}{LISA-13B$_{\!}$~\cite{lai2024lisa}}& \!\!\!\pub{CVPR2024}\!\!  & 45.2 & 47.9 & 46.6 & 34.3 & 39.1 & 36.7 & 39.8 & 43.5 & 41.6 & 8.6 \\
  \multicolumn{1}{r}{{VISA-7B}$_{\!}$~\cite{yan2024visa}}&  \!\!\!\pub{ECCV2024}\!\!  &49.2 & 52.6 & 50.9 & 40.6 & 45.4 & 43.0 & 44.9 & 49.0 & 46.9 & 15.5 \\
  \multicolumn{1}{r}{{VISA-13B}$_{\!}$~\cite{yan2024visa}}&  \!\!\!\pub{ECCV2024}\!\!  & 55.6 & 59.1 & 57.4 & 42.0 & 46.7 & 44.3 & 48.8 & 52.9 & 50.9 & 14.5 \\
   \multicolumn{1}{r}{{InstructSeg-3B}$_{\!}$~\cite{wei2024instructseg}}&  \!\!\!\pub{arXiv2025}\!\! & 54.8 & 59.2 & 57.0 & 49.2 & 54.7 & 51.9 & 52.0 & 56.9 & 54.5 & - \\
\multicolumn{1}{r}{{HyperSeg-3B}$_{\!}$~\cite{wei2024hyperseg}}&  \!\!\!\pub{arXiv2025}\!\! & 56.0 & 60.9 & 58.5 & 50.2 & 55.8 & 53.0 & 53.1 & 58.4 & 55.7 & - \\
\multicolumn{1}{r}{{GLUS-7B}$_{\!}$~\cite{lin2025glus}}&  \!\!\!\pub{CVPR2025}\!\! & 56.0 & 60.7 & 58.3 & 48.8 & 53.9 & 51.4 & 52.4 & 57.3 & 54.9 & 17.9 \\
   \multicolumn{1}{r}{{VRS-HQ-7B}$_{\!}$~\cite{gong2025devil}}&\!\!\!\pub{CVPR2025}\!\! & 59.8 & 64.5 & 62.1 & 53.5 & 58.7 & 56.1 & 56.6 & 61.6 & 59.1 & 19.7 \\
   \multicolumn{1}{r}{{VRS-HQ-13B}$_{\!}$~\cite{gong2025devil}}&\!\!\!\pub{CVPR2025}\!\!   & \textbf{61.1} & 65.5 & \underline{63.3} & \underline{54.1} & 59.4 & \underline{56.8} & \underline{57.6} & 62.5 & \underline{60.0} & 18.9\\
   \hline
   \rowcolor{lightgray}
    \multicolumn{1}{r}{\textbf{Veason-R1-3B}}&\!\!\!\pub{Ours}\!\!  & 60.3 & \underline{65.6} & 63.0 & 53.6 & \underline{60.0} & \underline{56.8} & 56.9 & \underline{62.8} & 59.9 & \textbf{28.5}\\
    \rowcolor{lightgray}\multicolumn{1}{r}{\textbf{Veason-R1-7B}}&\!\!\!\pub{Ours}\!\!  & \underline{60.7} & \textbf{66.5} & \textbf{63.6} & \textbf{55.8} & \textbf{62.2} & \textbf{59.0} & \textbf{58.2} & \textbf{64.4} & \textbf{61.3} & \underline{27.0} \\
\shline
\end{tabular}
}
\end{table*}

%% file: Tables/reason_mevis_results_.tex
\begin{table*}[!tbp]
\begin{minipage}{0.49\textwidth}
\captionsetup{width=0.9\linewidth}
\caption{
\textbf{Benchmark Comparison on the ReasonVOS.}}
\label{ReasonVOS}
\footnotesize
\vspace{-1mm}
\scalebox{0.9}{
\begin{tabular}{cl||cccc}
\shline
\ \ \ \ \ \ \ \ \ \ \ \ Methods & & $\mathcal{J}$ & $\mathcal{F}$ & $\mathcal{J}\&\mathcal{F}$ \\
\shline
\multicolumn{1}{r}{OnlineRefer$_{\!}$~\cite{wu2023onlinerefer}}&  \!\!\!\pub{ICCV2023}\!\! & 34.6 & 42.9 & 38.7 \\
\multicolumn{1}{r}{SgMg$_{\!}$~\cite{miao2023spectrum}}&  \!\!\!\pub{ICCV2023}\!\! & 33.7 & 38.7 & 36.2 \\
     \multicolumn{1}{r}{LISA-7B$_{\!}$~\cite{lai2024lisa}}&  \!\!\!\pub{CVPR2024}\!\! & 29.1 & 33.1 & 31.1 \\
     \multicolumn{1}{r}{VideoLISA-3.8B$_{\!}$~\cite{bai2024one}}&  \!\!\!\pub{NeurIPS2024}\!\! & 45.1 & 49.9 & 47.5  \\
     \multicolumn{1}{r}{GLUS-7B$_{\!}$~\cite{lin2025glus}}&  \!\!\!\pub{CVPR2025}\!\! & 47.5 & 52.4 & 49.9  \\
\hline
  \rowcolor{lightgray}  \multicolumn{1}{r}{\textbf{Veason-R1-3B}}&\!\!\!\pub{Ours}\!\!  & \underline{51.8} &	\underline{58.5} &	\underline{55.2}\\
  \rowcolor{lightgray}  \multicolumn{1}{r}{\textbf{Veason-R1-7B}}&\!\!\!\pub{Ours}\!\!  & \textbf{56.0} & \textbf{63.8} & \textbf{59.9} \\
\shline
\end{tabular}
}
\hfill
\end{minipage}
\begin{minipage}{0.5\textwidth}
\caption{
\textbf{Benchmark Comparison on the MeViS. 
}}
\label{mevis}
\footnotesize
\vspace{-1mm}
\scalebox{0.9}{
\begin{tabular}{cc||ccc}
\shline
Methods &  & $\mathcal{J}$ & $\mathcal{F}$ & $\mathcal{J}\&\mathcal{F}$ \\
\shline
\multicolumn{1}{r}{LMPM$_{\!}$~\citep{ding2023mevis}} & \!\!\!\pub{ICCV2023}\!\! & 34.2 & 40.2 & 37.2 \\
\multicolumn{1}{r}{{VISA-13B}$_{\!}$~\citep{yan2024visa}} & \!\!\!\pub{ECCV2024}\!\! & 41.8 & 47.1 & 44.5 \\
\multicolumn{1}{r}{VideoLISA-3.8B$_{\!}$~\citep{bai2024one}} & \!\!\!\pub{NeurIPS2024}\!\! & 41.3 & 47.6 & 44.4 \\
\multicolumn{1}{r}{GLUS-7B$_{\!}$~\cite{lin2025glus}}&  \!\!\!\pub{CVPR2025}\!\! & \underline{48.5} & \underline{54.2} & \underline{51.3}  \\
\multicolumn{1}{r}{{VRS-HQ-13B}$_{\!}$~\cite{gong2025devil}}&\!\!\!\pub{CVPR2025}\!\!  & 48.0 & 53.7 & 50.9 \\
\hline
  \rowcolor{lightgray}  \multicolumn{1}{r}{\textbf{Veason-R1-3B}} & \!\!\!\pub{Ours}\!\! & 48.2 &	\underline{54.2} &	51.2\\
  \rowcolor{lightgray}  \multicolumn{1}{r}{\textbf{Veason-R1-7B}} & \!\!\!\pub{Ours}\!\! & \textbf{48.4} & \textbf{56.0} & \textbf{52.2} \\
\shline
\end{tabular}
}
\end{minipage}
\vspace{-1mm}
\end{table*}

%% file: Tables/reward_cotsft_ablation.tex
\begin{table}[t!]
\begin{minipage}{0.5\textwidth}
\caption{\textbf{Ablation Study of Reward Functions. } Combining\\
all reward components achieves the highest performance. }
\label{reward ablation}
\footnotesize
\vspace{-1mm}
\scalebox{1.0}{
\begin{tabular}{c|ccc|ccc}
\shline
  & \multicolumn{3}{c|}{Referring} & \multicolumn{3}{c}{Reasoning} \\
  \multirow{-2}{*}{Reward} & $\mathcal{J}$ & $\mathcal{F}$ & $\mathcal{J}\&\mathcal{F}$ &$\mathcal{J}$ & $\mathcal{F}$ & $\mathcal{J}\&\mathcal{F}$\\
\shline
     w/o $\mathcal{R}_{k}$ & 58.8 & 64.4 & 61.6 & 51.9 & 57.9 &  54.9 \\
     w/o $\mathcal{R}_{s}$ & 58.1 & 63.9 & 61.0 & 50.3 & 57.5 & 53.9 \\
     w/o $\mathcal{R}_{u}$ & \underline{59.6} & \underline{65.3} & \underline{62.4} & \underline{53.0} & \underline{59.3} & \underline{56.1} \\
     \hline
     \rowcolor{lightgray} \textbf{Ours} &  \textbf{60.3} & \textbf{65.6} & \textbf{63.0} & \textbf{53.6} & \textbf{60.0} & \textbf{56.8} \\
\shline
\end{tabular}
}

\end{minipage}
\begin{minipage}{0.45\textwidth}
\vspace{3mm}
\caption{\textbf{Ablation Study of Training Strategy. } The integration of CoT-SFT and GRPO achieves the best result.}
\label{cotsft ablation}
\footnotesize
\vspace{-1mm}
\scalebox{0.9}{
\begin{tabular}{c||ccc|ccc}
\shline
  & \multicolumn{3}{c|}{Referring} & \multicolumn{3}{c}{Reasoning} \\
 \multirow{-2}{*}{Strategy} & $\mathcal{J}$ & $\mathcal{F}$ & $\mathcal{J}\&\mathcal{F}$ &$\mathcal{J}$ & $\mathcal{F}$ & $\mathcal{J}\&\mathcal{F}$\\
\shline
     Qwen2.5-VL & 19.1 & 22.8 & 20.9 & 17.3 & 21.7 & 19.5 \\
     \hline
     SFT & 36.4 & 40.2 & 38.3 & 31.0 & 35.3 & 33.1 \\
     CoT-SFT & 48.1 & 54.1 & 51.1 & 37.9 & 45.3 & 41.6 \\
     Pure GRPO & 57.0 & 63.0 & 60.0 & 50.8 & 57.6 & 54.2 \\
     SFT+GRPO & \underline{58.9} & \underline{64.4} & \underline{61.6} & \underline{52.8} & \underline{59.2} & \underline{56.0} \\
     \hline
     \rowcolor{lightgray} \textbf{CoT-SFT+GRPO} &  \textbf{60.3} & \textbf{65.6} & \textbf{63.0} & \textbf{53.6} & \textbf{60.0} & \textbf{56.8} \\

\shline
\end{tabular}
}
\end{minipage}
\vspace{-3mm}
\end{table}

%% file: Tables/keyframe_ablation.tex
\begin{table}[tbp]
\caption{\textbf{Ablation Study on Joint Keyframe-grounding Training. } Joint training of keyframe selection and spatial grounding results in best performance.}
\label{keyframe ablation}
\footnotesize
\vspace{-1mm}
\scalebox{0.95}{
\begin{tabular}{c||ccc|ccc}
\shline
    & \multicolumn{3}{c|}{Referring} & \multicolumn{3}{c}{Reasoning} \\
  \multirow{-2}{*}{Strategy} & $\mathcal{J}$ & $\mathcal{F}$ & $\mathcal{J}\&\mathcal{F}$ &$\mathcal{J}$ & $\mathcal{F}$ & $\mathcal{J}\&\mathcal{F}$\\
\shline
     Grounding-only & \underline{55.8} & \underline{60.9} & \underline{58.3} & \underline{48.2} & 54.4 & \underline{51.3} \\
     SFT-keyframe & 49.7 & 59.4 & 54.6 & 46.5 & \underline{55.6} & 51.0 \\
\hline
     \rowcolor{lightgray} \textbf{Ours} &  \textbf{60.3} & \textbf{65.6} & \textbf{63.0} & \textbf{53.6} & \textbf{60.0} & \textbf{56.8} \\
\shline
\end{tabular}
}
\vspace{-1mm}
\end{table}